\documentclass[12pt,journal,compsoc,onecolumn]{IEEEtran}
\usepackage{amsmath,epsfig}
 \usepackage{lingmacros}
 \usepackage{tree-dvips}
\usepackage{amsmath} 
\usepackage{amssymb}
\usepackage{pifont}
\usepackage[noadjust]{cite}

\newlength{\bibitemsep}
\newlength{\bibparskip}
\let\oldthebibliography\thebibliography
\renewcommand\thebibliography[1]{
  \oldthebibliography{#1}
  \setlength{\parskip}{\bibitemsep}
  \setlength{\itemsep}{\bibparskip}
}

\def\x{{\mathbf x}}

\def\XX{{\bf X}}

\def\Y{{\cal Y}}
\def\S{{\cal I}}

\def\x{{\bf x}}
\def\y{{\bf y}}
\def\D{{\cal D}}

\def\F{{ f}}
\def\DD{{d}}
\def\tr{{\bf tr}}
\def\1{{\bf 1}}
\def\I{{\cal I}} 
\def\p{{p}}
\def\q{{q}}
\def\x{{\bf x}}  
\def\Y{{\cal  Y}}  
\def\V{{\bf D}}
\newtheorem{proposition}{Proposition}

\usepackage[ruled,vlined]{algorithm2e}

\title{Frugal Satellite Image Change Detection with Deep-Net Inversion} 
\author{Hichem Sahbi \ \ \ \ \ \  \ \ \ \ \ \  \ \ \ \ \ \   \ \ \ \ \ \ Sebastien Deschamps\\
$ $ \\
 Sorbonne University, CNRS, LIP6,  F-75005, Paris, France  \\
}

\begin{document}
 \maketitle
\begin{abstract}

Change detection in satellite imagery seeks to find occurrences of targeted changes in a given scene taken at different instants. This task has several applications ranging from land-cover mapping, to anthropogenic activity monitory as well as  climate change and natural hazard damage assessment.  However, change detection is highly challenging due to the acquisition conditions and also to the subjectivity of changes.\\
In this paper, we devise a novel algorithm for change detection based on active learning. The proposed method is based on a question \& answer model that probes  an oracle (user) about the relevance of changes only on a small set of critical images (referred to as virtual exemplars), and according to oracle's responses updates deep neural network (DNN) classifiers.  The main contribution resides in a novel adversarial model that allows learning the most representative, diverse and uncertain virtual exemplars (as inverted preimages of the trained DNNs) that challenge (the most) the trained DNNs, and this leads to a better re-estimate of these networks in the subsequent iterations of active learning.  Experiments show the out-performance of our proposed deep-net inversion against the related work. 
\end{abstract}

\section{Introduction}
\label{sec:intro}
 
 \noindent Satellite image change detection aims at localizing instances of targeted (relevant) changes in scenes acquired at different instants. The wide interest of this task ranges from anthropogenic activity monitoring, to phenology mapping, through natural hazard damage assessment~\cite{ref4,ref5}. This task is challenging as satellite images are subject to many pervasive changes resulting from sensors and acquisition conditions (occlusions, radiometric variations and shadows,  weather conditions as well as scene content). Early change detection work relies on comparisons of multi-temporal series, via image differences and thresholding, vegetation indices, principal component and change vector analyses \cite{ref7,ref9,ref11,ref13}. Other work either (i) requires a preliminary step that mitigates the effect of pervasive changes using different normalization techniques including shadow removal, radiometric correction and by estimating the parameters of sensors for registration,  etc.  \cite{ref14,ref15,ref17,ref20,refffabc8}, or  (ii) considers these irrelevant variations as a part of scene appearance modeling using statistical machine learning~\cite{ref21,ref25,ref26,ref27,ref28,refffabc5,refffabc6,refffabc7,ICPR2016sahbi}. \\
 
\noindent The success of  machine learning models in particular, including deep neural networks, is highly reliant on the availability of large collections of hand-labeled reference images that capture the inherent variability of relevant and irrelevant changes as well as the user's targeted relevant changes~\cite{refffabc0,refffabc1}. However, in practice, labeled data are scarce and their hand-labeling is time and effort demanding, and even when available they suffer from domain-shift as their labeling may not reflect the user's subjectivity and intention. Several solutions seek to make machine learning methods frugal and less labeled-data hungry \cite{refffabc3,refffabc4} including few shot \cite{reff45} and self-supervised  learning \cite{refff2}; however,  these methods are agnostic  to the users' intention. Hence, solutions based on active learning  \cite{reff1,reff2,reff16,reff15,reff53,reff12,reff74,reff58,reff13} are rather more appropriate and consist in {\it frugally probing} the user (a.k.a. oracle) about the relevance of observed changes, and according to the oracle's responses, train decision  criteria that best suit the user's intention and scene acquisition conditions. \\

\indent In this paper, we devise a new satellite image change detection algorithm based on frugal training of deep neural networks.  Starting from a large unlabeled dataset, the proposed model is interactive and consists in querying the oracle about the labels of {\it critical} data whose positive impact on the trained deep networks is the most important. These critical data (also dubbed as virtual exemplars) are obtained by {\it deep-net inversion} using an adversarial loss that allows synthesizing representative and diverse exemplars whose classification scores are also ambiguous. These virtual exemplars are learned -- instead of being sampled from a {\it fixed} pool of unlabeled data -- in order to challenge (the most) the previously trained change detection criteria leading to a better re-estimation of these criteria at the subsequent iterations of change detection. Note that the formulation presented in this paper, while being adversarial, is conceptually different from the ones widely used in generative adversarial networks (GANs) \cite{refffabc}; indeed, whereas GANs seek to generate {\it fake} exemplars that mislead the trained discriminators, our formulation aims instead at generating {\it critical} exemplars --- for further annotations --- whose impact on the subsequent learned deep network classifiers is the most noticeable. Differently put, the proposed framework allows to sparingly probe the oracle only on the most representative, diverse and uncertain exemplars that challenge the current deep network discriminator, and eventually lead to more accurate ones in the following iterations of change detection.  Extensive change detection experiments corroborate these claims and show the effectiveness of our {\it deep-net inversion and  exemplar learning}  models against comparative methods. 

\section{Proposed method}
\label{sec:format}
Let $\I_r = \{\p_1, \dots , \p_n\}$, $\I_t = \{\q_1, \dots , \q_n\}$  denote two registered satellite images captured at two instants $t_0$, $t_1$, with $\p_i$, $\q_i \in \mathbb{R}^d$. Considering  $\I = \{\x_1,\dots, \x_n\}$, with each $\x_i=(\p_i, \q_i)$, and $\Y = \{\y_1, \dots, \y_n\}$ the underlying unknown labels; our goal is to train a classifier $f$ --- as a deep convolutional network --- that predicts the unknown labels in $\{\y_i\}_i$ with $\y_i = 1$ if the patch $\q_i$  corresponds to a ``change'' w.r.t. the underlying patch $\p_i$, and $\y_i = 0$ otherwise. Training $f$ requires data hand-labeled by an oracle. Our goal is to make the design of $f$ {\it label-frugal} and as accurate as possible.
\subsection{Interactive change detection at a glance}
In order to design our change detection algorithm, we consider a question \& answer interactive process which first provides the oracle with very few {\it critical patch pairs} for labeling, and then updates change detection criteria accordingly. In what follows, the subset of critical data, at any iteration $t$, is referred to as {\it display}, and denoted as  $\D_t \subset \I$, and $\Y_t$ its unknown labels. Starting from a random display $\D_0$, we train change detection criteria $f_0,\dots,f_{T-1}$ iteratively according to the subsequent steps\\

\noindent 1/ Obtain $\Y_t$ from the oracle, and train $f_t (.)$ on $\cup_{k=0}^t (\D_k,\Y_k)$ using graph convolution networks (GCNs)~\cite{sahbi2021b}.\\

2/ Find the next display $\D_{t+1}\subset \S\backslash\cup_{k=0}^t \D_k$. It is clear that a strategy that considers all possible displays $\D \subset \S\backslash\cup_{k=0}^t \D_k$, trains the underlying criteria  $f_{t+1} (.)$ on $\D \cup_{k=0}^t \D_k$,  and keeps the display  with the highest performances is highly combinatorial. Furthermore, it requires labeling each of these displays and this is clearly intractable. In this paper, we rely on active learning display selection strategies, which are rather more appropriate; nevertheless, one should be cautious about these strategies as many of them are shown to be equivalent to (or even worse than) basic strategies that pick data uniformly randomly (see for instance \cite{reff2} and references within). \\

As the main contribution of this work, our proposed display selection strategy  is different from common ones (see e.g. \cite{refff33333}) and relies on hallucinating virtual exemplars instead of taking them from existing pools of data. This design allows exploring {\it more flexibly} the uncharted parts of unlabeled data. These exemplars are selected as the most diverse, representative and uncertain data  so they {\it challenge the most} the current change detection criteria, leading to better re-estimate of these criteria in the subsequent iterations of active learning. 

\subsection{Virtual exemplars: column-stochastic model}\label{formulation}
As obtaining labels is usually highly expensive, we train a deep network $f_{t+1}(.)$ on ${\cal D}_{t+1} \cup \dots \cup {\cal D}_{0}$ where ${\cal D}_{t+1}$ is a subset of critical virtual exemplars (denoted as $\{\V_k\}_{k=1}^K$) whose labeling is frugally obtained from the oracle. In contrast to our previous work~\cite{refsahbiigarss22}, we consider a variant which defines for each virtual exemplar $\V_k$ a distribution $\{\mu_{ik}\}_{i=1}^n$  that measures the conditional probability of associating $\V_k$ to the $n$-training samples. With this variant, the virtual exemplars together with their distributions  $\{\mu_{ik}\}_{i,k}$ are obtained by minimizing  the following surrogate problem

{
\begin{equation}\label{eq01}
\hspace{-0.25cm}\begin{array}{ll}
\displaystyle   \min_{\V; \mu \in \Omega_c}    & \displaystyle   \tr\big (\DD(\V,\XX) \ \mu \big)  \  + \ \alpha  \ \big[\frac{1}{K} \ \mu \ \1_K \big]^\top \log \big[\frac{1}{K}  \mu \ \1_K\big] \\
                &  \  +  \  \beta \ \tr\big(\F(\V)^\top \  \log \F(\V)\big) \ + \ \gamma \ \tr(\mu \log \mu^\top), 
\end{array}
\end{equation}}

\noindent  here   $\Omega_c=\{\mu :  \mu \geq 0; \mu^\top \1_n  = \1_K\}$ guarantees the column-stochasticity of the memberships $\{\mu_{ik}\}_i$ instead of row-stochasticity in~\cite{refsahbiigarss22}, and   $\1_{K}$, $\1_{n}$ denote two vectors of $K$ and $n$ ones respectively. In Eq.~\ref{eq01} the symbol $^\top$ stands for  the matrix transpose operator, $\V \in \mathbb{R}^{d \times K}$  is a matrix whose k-th column represents the k-th learned virtual exemplar, $\mu \in \mathbb{R}^{n \times K}$ is a learned matrix whose k-th column corresponds to the conditional probability of assigning $\V_k$ to each of the $n$ input data, and $\log$ is applied entrywise. In Eq.~\ref{eq01}, $\DD(\V,\XX) \in \mathbb{R}^{K \times n}$ is the matrix of the euclidean distances between input data in $\XX$ and the virtual exemplars in $\V$ whereas $\F(\V) \in \mathbb{R}^{2 \times K} $ is the softmax output of the learned network $f$. The first term in Eq.~\ref{eq01} measures how representative are $\{\V_{k}\}_k$ w.r.t. the unlabeled samples in $\I$. This term, as defined, has a major advantage compared to \cite{refsahbiigarss22}; indeed, due to column-stochasticity of $\mu$, the first term in Eq.~\ref{eq01} reaches (more easily) smaller values than its counterpart in \cite{refsahbiigarss22} (whose $\mu$ is row-stochastic),  and this allows producing virtual exemplars closer to the input data and thereby inheriting more accurate labels when the oracle annotates their closest input data. 

The second term in Eq.~\ref{eq01} also captures diversity, not in the virtual exemplars themselves (as in \cite{refsahbiigarss22}), but instead in  {\it how training data polarize (or attract) the virtual exemplars}; this term reaches its minimum when all the training samples evenly attract the virtual data.  The third and fourth terms are similar to \cite{refsahbiigarss22}. The former measures the {\it ambiguity} (or uncertainty) in $\V$ as the entropy of the scoring function (it reaches its smallest value when virtual exemplars in  $\V$ are evenly scored w.r.t different classes) while the latter acts as a regularizer which favors uniform conditional probability distribution  $\{\mu_{ik}\}_i$. Finally, all these terms are mixed using the coefficients $\alpha, \beta, \gamma \geq 0$.

\subsection{Optimization}
In order to optimize Eq.~\ref{eq01}, we consider, for each change detection cycle $t$,  an EM-like  procedure that first fixes the display $\V$ and solves the optimization problem w.r.t. $\mu$ as shown in the following proposition.

\begin{proposition}
The optimality conditions of Eq.~(\ref{eq01}) lead  to
{
\begin{equation}\label{eq2}
\begin{array}{lll}
 \mu^{(\tau+1)}& :=&\displaystyle  \hat{\mu}^{(\tau+1)} \  \textrm{\bf diag} \big( \1^\top_n \hat{\mu}^{(\tau+1)}\big)^{-1}, \\
\end{array}  
\end{equation}}
being  $\hat{\mu}^{(\tau+1)}$
{ 
\begin{equation}\label{eq3}
\begin{array}{l}
\exp\bigg\{-\frac{1}{\gamma}[\DD(\XX,\V^{(\tau)}) + \frac{\alpha}{K} ( \1_n +  \log \frac{1}{K}\mu^{(\tau)} \1_K ) \1_K^\top ]\bigg\},
\end{array} 
  \end{equation} }
  \noindent here  $\textrm{\bf diag}(.)$ maps a vector  to a diagonal matrix.                        
\end{proposition} 

Due to space limitation, details of the proof are omitted and  result from the optimality conditions of Eq.~\ref{eq01}'s gradients. Once $\mu$ optimized (and fixed) the problem is solved w.r.t. $\V$ using stochastic gradient descent. A loss  ${\cal L}(\V)$ is composed of the first and the third term  of Eq.~\ref{eq01}, then its gradient (w.r.t. $\V$)  is back-propagated through the GCN layers till the input (virtual data) of the learned GCNs using the chain rule. Note that ${\mu}^{(0)}$ and  ${\V}^{(0)}$ are initially set to random values and, in practice, the procedure converges to an optimal solution (denoted as $\tilde{\mu}$, $\tilde{\V}$) in few iterations. This solution defines the most {\it relevant} virtual exemplars of $\D_{t+1}$ (according to criteria in Eq. \ref{eq01}) which are used to train the subsequent classifier $f_{t+1}$. 

\section{Experiments}

Change detection experiments  are conducted on the Jefferson dataset. The latter includes $2,200$ non-overlapping aligned patch pairs (of $30\times 30$ RGB pixels each) taken from bi-temporal GeoEye-1 satellite images of $2, 400 \times  1, 652$ pixels with  a spatial resolution of 1.65m/pixel.  These images have been  taken from the area of Jefferson (Alabama) in 2010 and 2011 and correspond to many changes (building destruction, etc.)  due to tornadoes as well as no-changes (including irrelevant ones as clouds, occlusions, etc).  The ground-truth associated to this dataset includes 2,161 negative pairs (no changes and irrelevant ones) and only 39 positive pairs (relevant changes), so more than $98\%$ of these data correspond to irrelevant changes, and this makes localizing changes very challenging.  In all our experiments,  we split the whole dataset evenly; (i) for training (our display and learning models) and (ii) for testing performances.  As the two classes (changes/no-changes) are highly imbalanced, we measure performances using the equal error rate (EER) for different sampling percentages. At each iteration $t$,  the sampling percentage corresponds to  $(\sum_{k=0}^{t-1} |\D_k|/(|\I|/2))\times 100$  with $|\I|=2,200$ and $|\D_k|$ set to $16$.  Smaller EER implies better performances.\\ 
\begin{figure}[tbp]
\center
\includegraphics[angle=0,width=0.5\linewidth]{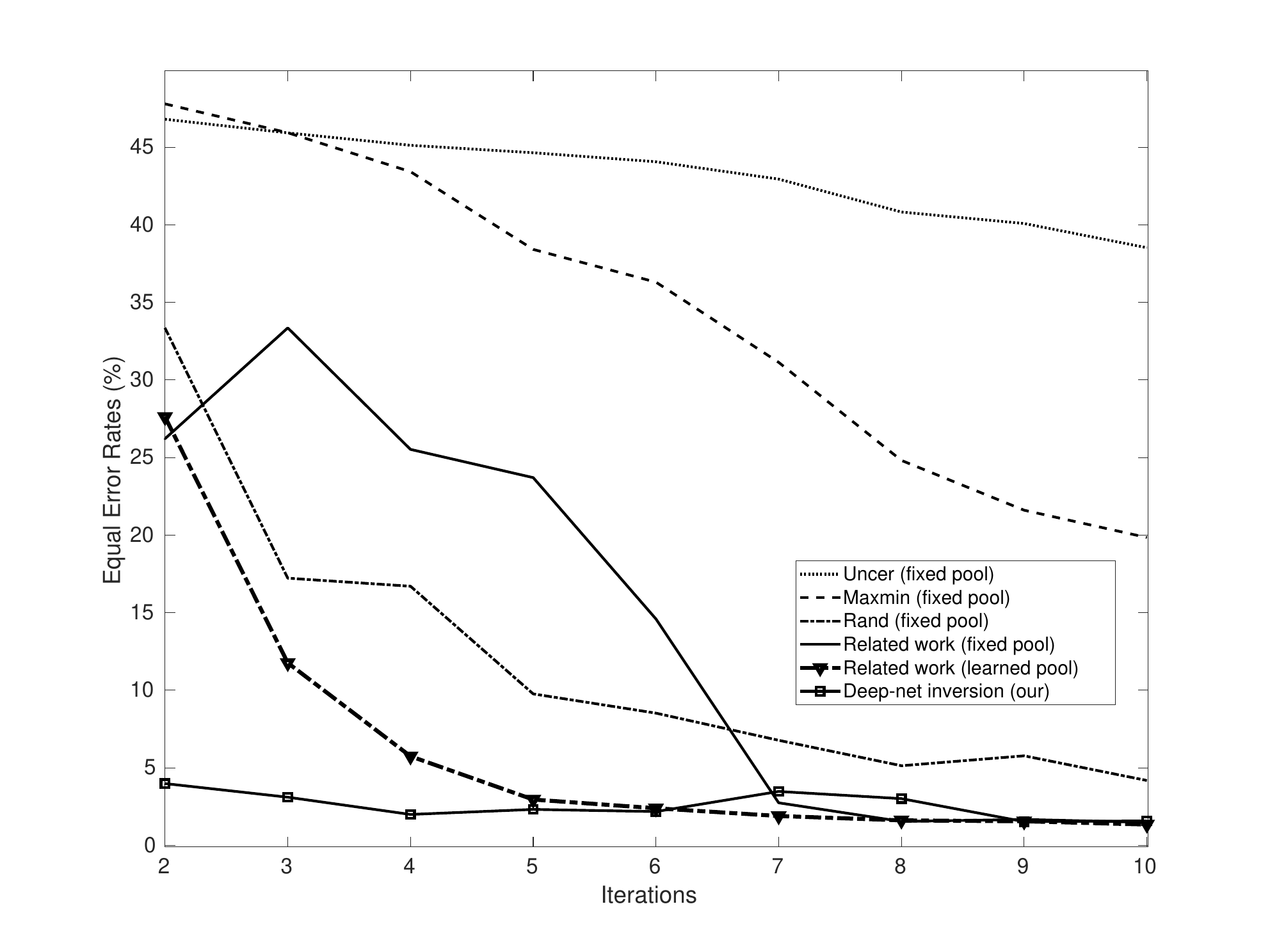}
\caption{Comparison of different sampling strategies w.r.t. different iterations and the underlying sampling rates in figure~\ref{tab3}. Here Uncer and Rand stand for uncertainty and random sampling respectively. Note that fully-supervised learning achieves an EER of $0.94 \%$.  Related works (fixed and learned pools) stand respectively for the methods in \cite{refff33333} and \cite{refsahbiigarss22}.}\label{tab2}
\end{figure}

\noindent {\bf Performances \& comparison.} We compare our proposed method against other related sampling strategies including fixed-pool (random search, maxmin, uncertainty ...) \cite{refff33333} as well as learned-pool methods~\cite{refsahbiigarss22}. Random consists in selecting samples from the pool of unlabeled data while uncertainty consists in collecting, from the same pool, the display whose classifier scores are the most ambiguous (i.e., similar across different classes). Maxmin greedily samples data in $\D_{t+1}$; each sample in  $\x_i \in \D_{t+1} \subset \S\backslash \cup_{k=0}^t \D_k$ is taken by {\it maximizing its minimum distance w.r.t.  $\cup_{k=0}^t \D_k$},  leading to the most different samples in $\D_{t+1}$. We further compare our display model against \cite{refff33333} which consists in assigning marginal membership probabilities  (instead of conditional ones) to the whole unlabeled set and selecting the display with the highest membership values. We also show the performance of virtual exemplar learning model in \cite{refsahbiigarss22} which relies on support vector machines (SVMs) and row-stochastic formulation as already discussed in section~\ref{formulation}.  Finally, we show a lower bound on EER performances using a fully supervised setting which builds a unique classifier on top of the full training set whose annotation is taken from the ground-truth. The EER performances of all these settings are shown in Figure~\ref{tab2} through different iterations and sampling rates,  as well as the aforementioned display selection strategies. From the EERs, most of the comparative methods are powerless to spot the change class sufficiently well; indeed, random and maxmim capture diversity during early change detection iterations but they are less effective later, and uncertainty lacks diversity. The comparative fixed-pool work~\cite{refff33333} (which mixes diversity, representativity and uncertainty on a fixed pool) captures diversity and refines better change detections, however, it suffers from the rigidity of the sampled data (especially at the early iterations), while the comparative learned-pool method~\cite{refsahbiigarss22} (which mixes the same terms, and learns pools of data with SVMs and row-stochastic memberships) is more flexible, but it is less effective than our proposed deep-net inversion method particularly at highly frugal regimes.

\begin{figure}[tbp]
\center
\includegraphics[angle=0,width=0.5\linewidth]{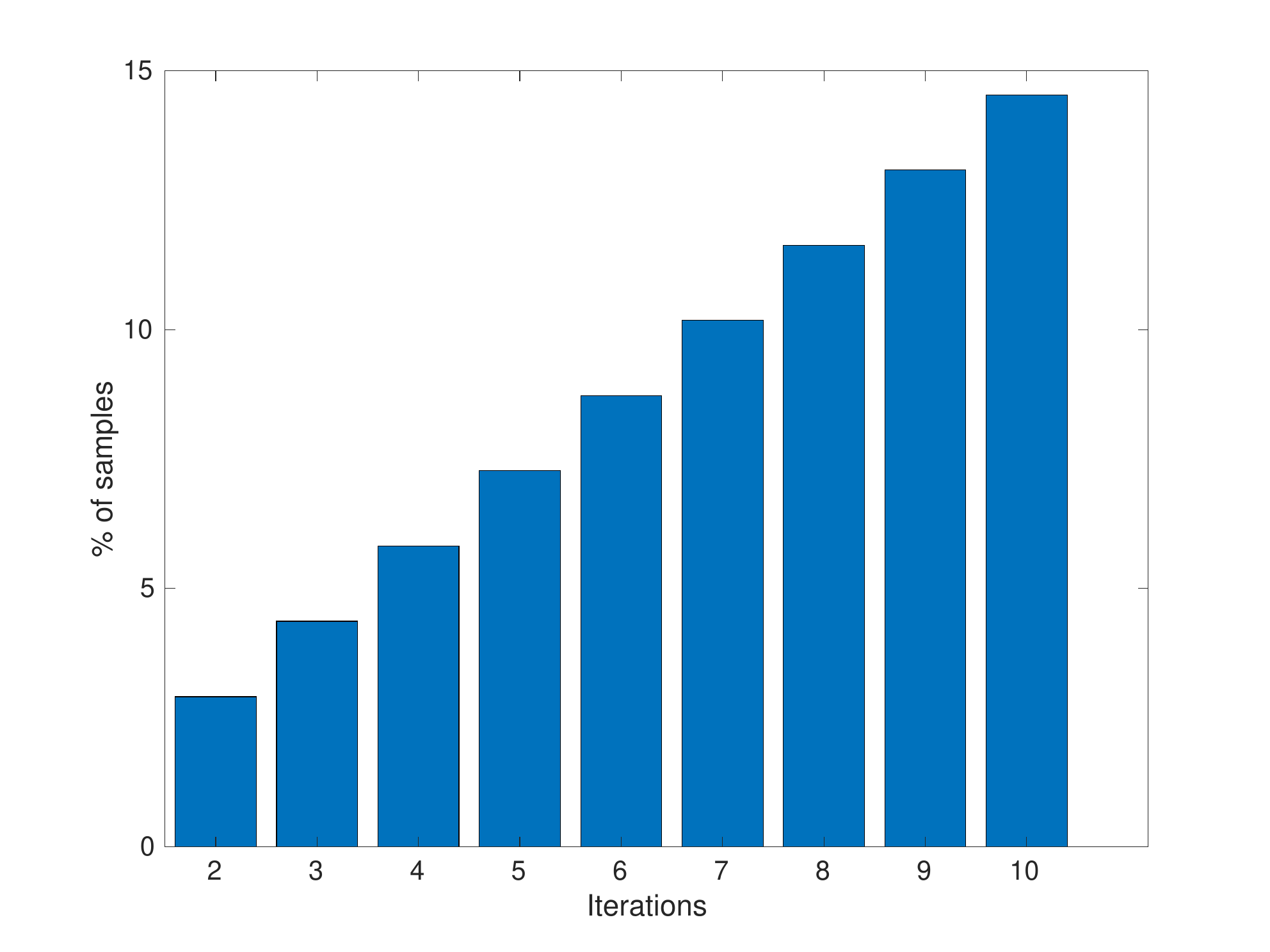}
\caption{This figure shows sample rates for each iteration of active learning.}\label{tab3}
\end{figure}

\section{Conclusion}
We introduce in this paper a satellite image change detection algorithm based on active learning. The particularity of our contribution resides in a new adversarial learning model based on deep neural network (DNN) inversion. The latter allows to flexibly train small sets of critical data (dubbed as virtual exemplar) whose labels are frugally obtained from the oracle. These virtual exemplars correspond to the most diverse, representative and uncertain data that challenge (the most) the current DNNs leading to better re-estimate of these DNNs at the subsequent iterations of active learning.  Extensive experiments show the out-performance of our virtual exemplar learning model against different baselines, including fixed and learned pools, as well as the related work. 

\newpage 
{

}

\end{document}